\documentclass[conference]{IEEEtran}
\usepackage{times}

\usepackage{cuted}
\usepackage{capt-of}
\usepackage[pdftex]{graphicx}   
\let\labelindent\relax
\usepackage{enumitem}
\usepackage{graphics}
\usepackage{wrapfig}
\usepackage{float}
\usepackage{svg}
\usepackage[font=footnotesize]{caption}
\usepackage{subcaption}

\usepackage{amsmath}
\usepackage{amssymb}
\usepackage{amsfonts}          
\usepackage{amsthm}
\usepackage{mathtools}         
\usepackage{bm}

\usepackage{booktabs}
\usepackage{multirow}

\usepackage{algorithm}
\usepackage{algorithmic}       

\usepackage{xcolor}            
\usepackage{color}             
\usepackage{mdframed}
\usepackage{tcolorbox}
\usepackage{listings}

\usepackage{multicol}
\usepackage{comment}
\usepackage{xparse}
\usepackage{textcomp}
\usepackage{epigraph}
\usepackage{url}

\usepackage[numbers,square,comma,sort&compress]{natbib}
\usepackage[bookmarks=true,pagebackref=true]{hyperref}
\hypersetup{
    colorlinks=true,
}
\usepackage{cleveref}          

\def \FullMethodName {Reinforcement Learning from RL Token}
\def \MethodName {\textsc{RLT}}

\title{RL Token: Bootstrapping Online RL with Vision-Language-Action Models}

\author{
  \IEEEauthorblockN{
    Charles Xu,
    Jost Tobias Springenberg,
    Michael Equi,
    Ali Amin,
    Adnan Esmail,
    Sergey Levine,
    Liyiming Ke
  }
  \IEEEauthorblockA{\textbf{Physical Intelligence}\\
 \url{https://pi.website/research/rlt}
  }
}

\begin{document}
\maketitle
\begin{strip}
\includegraphics[width=\textwidth]{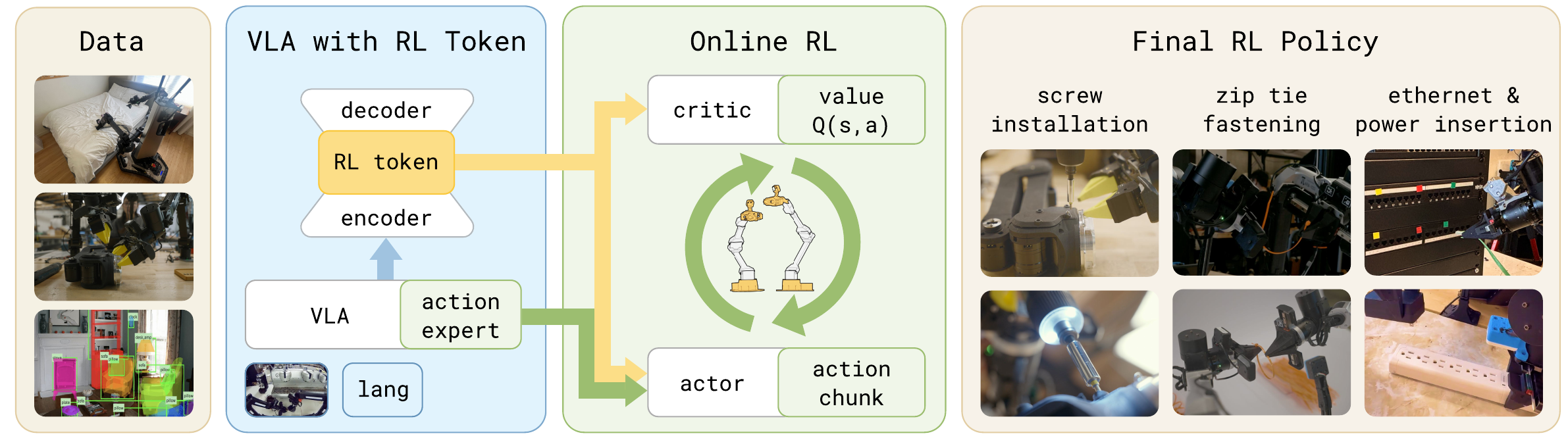}
    \captionof{figure}{Our method introduces an ``RL token'' into the VLA by training an encoder and decoder to produce a compact and meaningful representation from a VLA's internal features. The extracted representation is then used to train lightweight actor-critic networks with sample-efficient online RL, enabling very precise tasks to be fine-tuned with a few hours or even minutes of robot experience.}
    \label{fig:overview}
\end{strip}

\begin{abstract}
Vision-language-action (VLA) models can learn to perform diverse manipulation skills ``out of the box,'' but achieving the precision and speed that real-world tasks demand requires further fine-tuning -- for example, via reinforcement learning (RL).
We introduce a lightweight method that enables sample-efficient online RL fine-tuning of pretrained VLAs using just a few hours of real-world practice. We (1) adapt the VLA to expose an ``RL token,'' a compact readout representation that preserves task-relevant pretrained knowledge while serving as an efficient interface for online RL, and (2) train a small actor-critic head on this RL token to refine the actions, while anchoring the learned policy to the VLA. 
Online RL with the RL token (\MethodName{}) makes it possible to fine-tune even large VLAs with RL quickly and efficiently.
Across four real-robot tasks (screw installation, zip tie fastening, charger insertion, and Ethernet insertion) \MethodName{} improves the speed on the hardest part of the task by up to 3$\times$ and raises success rates significantly within minutes to a few hours of practice. It can even surpass the speed of human teleoperation on some of the tasks. 
\end{abstract}

\IEEEpeerreviewmaketitle

\section{Introduction}
\label{sec:introduction}

General-purpose vision-language-action (VLA) models can learn a wide range of diverse manipulation skills from data. Yet they often struggle in the last millimeter of execution: motions can be slow, successful completion may require pauses and retries, and small errors at critical stages of precise tasks can compound into failure. A natural way to address this challenge is to fine-tune VLAs with reinforcement learning (RL). By practicing on the target task, RL can improve precisely the stages of the task that are most critical for success, which are often the stages most sensitive to small errors and the hardest to cover reliably with demonstrations alone. But real-world robotics operates under a tight budget: every episode takes time, every failure consumes effort and wear, and meaningful adaptation often has to happen within a few hours of practice.

Sample-efficient fine-tuning of VLAs, however, presents major challenges. On one hand, conventional methods for RL training of foundation models~\citep{Li2025SimpleVLA_RL,li2025grrlgoingdexterousprecise,pi06vla} rely on large-scale data and can be inefficient for rapid online adaptation. On the other hand, data-efficient real-world RL methods~\citep{luo2024precise,rl100} typically train much smaller models, which can improve within hours but sacrifice the generalization capabilities of the VLA. The central question, then, is how to leverage the VLA’s generalization while achieving the speed and sample efficiency of lightweight online RL.

We introduce a practical recipe that bootstraps fast online reinforcement learning with representations obtained from a pretrained VLA policy. Our key idea is to adapt the VLA so that it exposes a compact interface that can be used for sample-efficient online RL. We achieve this by training the VLA to expose an \emph{RL token}, a compressed representation that makes task-relevant pretrained knowledge accessible to a lightweight online RL policy. Running RL with this RL token (\MethodName{}) creates a simple division of labor: the frozen VLA provides broad perceptual understanding and action recommendations, while the lightweight actor and critic adapt the policy to succeed at the hardest parts of a task online. To make this practical in the sample-efficient real-world regime, our method uses a sample-efficient online RL algorithm to train the small actor and critic networks that use the RL token representation, with an additional regularizer to anchor the actor to the VLA action, so that online RL refines promising behaviors rather than learning from scratch.

We evaluate \MethodName{} on four challenging robot manipulation tasks that can require millimeter or sub-millimeter precision: screw installation, zip tie fastening, Ethernet, and charger insertion.
Across these tasks, \MethodName{} improves both success rate and execution speed within a few hours of online training. The largest gains appear in the critical phases of the tasks, which require high precision and determine task success, where \MethodName{} speeds up execution by up to 3$\times$ and substantially improves success rates, for example from 20\% to 65\% for a challenging screw-insertion task. On one of the most dexterous parts of our tasks, policies trained with our method can surpass expert teleoperation speed while maintaining reliability. These results suggest that combining a VLA model with lightweight online RL offers a practical path to high-performance manipulation without extensive task-specific engineering.

\section{Related Work}
\label{sec:related}
\textbf{Vision-language-action models.}
Behavioral cloning from large demonstration datasets has recently emerged as the dominant paradigm for training generalist robot manipulation policies (see e.g.~\citep{rt22023arxiv,kim2024openvla,black2024pi_0,geminirobotics,wu2023unleashing,nvidia2025gr00tn1openfoundation}).
Two critical ingredients that have enabled this success are action chunking~\cite{zhao2023learningfinegrainedbimanualmanipulation}, which predicts multiple actions for sequential open-loop execution, and using expressive output distributions, such as diffusion~\cite{chi2023diffusion} or autoregressive generation~\cite{rt22023arxiv}, that can capture the multimodality inherent in demonstration data.
A further advancement came from using large pretrained vision-language models as a backbone for language-conditioned generalist policies, yielding vision-language-action (VLA) models~\cite{rt22023arxiv,kim2024openvla}.
These models import large web-scale prior knowledge into closed-loop robot policies.
Recent work has combined VLA backbones with chunked action generation, either via diffusion~\cite{black2024pi_0} or autoregressive tokenization~\cite{pertsch2025fast,minivla}, achieving state-of-the-art generalist manipulation.
While these policies exhibit impressive generalization capabilities~\cite{pi05,geminirobotics}, their performance on any given task is ultimately bounded by the quality and coverage of the teleoperation data they were trained on---achieving reliable success on precision-critical tasks remains difficult when demonstrations themselves are noisy or inconsistent.
 
\textbf{Real-world reinforcement learning.}
Reinforcement learning offers a natural way to go beyond the performance ceiling of demonstration data: by practicing on a task, the agent can discover faster, more precise, or more robust strategies that were never demonstrated.
In practice, real-world RL for robotics operates under tight sample budgets, since every robot rollout costs time and wear.
Off-policy actor-critic methods (e.g.~\cite{haarnoja2018soft,lillicrap2015continuous,fujimoto2018td3,Abdolmaleki2018MaximumAP}) address this by reusing transitions stored in a replay buffer, and sample efficiency can be pushed further by increasing the update-to-data ratio~\cite{hussing2024dissectingdeeprlhigh}, though regularization may be necessary to avoid instability~\cite{chen2021randomized}.
Crucially, off-policy methods can also incorporate human demonstration data to bootstrap learning (e.g.~\cite{ball2023efficient}), combining the strengths of imitation and RL.
A growing body of work has developed practical recipes for deploying RL on physical robots, including autonomous data-collection pipelines~\cite{zhu2020ingredients}, efficient learning frameworks such as SERL~\cite{luo2024serl,luo2024precise} and RL$^{100}$~\cite{rl100}, and human-in-the-loop variants that allow operators to intervene and provide corrections during autonomous execution~\cite{luo2024precise}.
These systems have demonstrated that off-policy actor-critic methods, combined with demonstrations and human corrections, can solve contact-rich manipulation tasks within hours of robot time.
However, they typically train small policies from scratch atop standard pretrained visual encoders (e.g., ResNet), forgoing the rich behavioral priors available in modern VLA models.
\MethodName{} bridges this gap by using a frozen VLA as both the perceptual backbone and the behavioral prior for a lightweight online RL policy.
 
\textbf{RL fine-tuning of VLA models.}
A rapidly growing line of work studies how to improve a pretrained VLA via RL.
These approaches vary primarily in \emph{what} is updated and \emph{how} the RL signal is incorporated. 
At one end of the spectrum, several methods update the full VLA model.
RECAP~\cite{pi06vla} trains the entire $\pi^*_{0.6}$ model end-to-end using offline RL via advantage-conditioned policy extraction: a distributional value function estimates per-timestep advantages, and the VLA is trained on all collected data---demonstrations, autonomous rollouts, and human interventions---with an optimality indicator that upweights high-advantage actions.
By iterating between on-robot data collection and offline RL updates, RECAP more than doubles throughput on complex long-horizon tasks such as espresso making, laundry folding, and box assembly.
Other works apply proximal policy optimization (PPO) or variants thereof to VLA fine-tuning (e.g.~\cite{Ren2025DPPO,Chen2025piRL,Li2025SimpleVLA_RL}), though on-policy methods are difficult to extend to real-world RL in a sample-efficient and scalable fashion. 
At the other end, lightweight methods avoid updating the full VLA and instead train a small auxiliary module on top of the frozen model.
ConRFT~\cite{chen2025conrft} freezes the VLA encoder and fine-tunes an action head using a consistency-based training objective together with a learned binary reward classifier, but operates on single-step actions without chunking on short-horizon tasks.
Policy Decorator~\cite{pi_dec} learns a residual policy whose output is scaled by a hand-tuned hyperparameter and added to the frozen VLA's prediction, but has been demonstrated only in simulation with high sample requirements (on the order of millions of steps).
Probe-Learn-Distill (PLD)~\cite{xiao2025selfimprovingvisionlanguageactionmodelsdata} pre-trains a critic with Cal-QL~\cite{nakamoto2023cal} on base-policy rollouts and then learns a single-step residual policy on top of the frozen VLA, optionally distilling the result back into the VLA via supervised fine-tuning.
GR-RL~\cite{li2025grrlgoingdexterousprecise} takes a multi-stage approach for specializing a generalist VLA on a long-horizon shoe-lacing task: it first performs offline filtered BC, and then performs online RL by learning a noise predictor that steers the frozen VLA's diffusion process in latent space~\citep{Wagenmaker2025DSRL}.
DSRL~\cite{Wagenmaker2025DSRL} similarly operates in the diffusion noise space, learning a latent policy that modulates the denoising process to steer actions toward high-return regions.

\MethodName{} shares with these methods the goal of improving a pretrained VLA without the cost of full-model RL, but differs in several key design choices.
First, \MethodName{} introduces an \emph{RL token}---a compact readout representation trained to compress the VLA's internal embeddings---that serves as the state observation for a lightweight actor-critic, preserving the VLA's pretrained perceptual structure while enabling efficient online learning.
Second, \MethodName{} operates over \emph{chunked actions} aligned with the VLA's native action interface, shortening the effective decision horizon for temporal-difference learning under sparse rewards at high control frequencies---in contrast to single-step methods~\cite{chen2025conrft,pi_dec,xiao2025selfimprovingvisionlanguageactionmodelsdata} that face a much longer credit-assignment problem.
Third, rather than predicting residuals or latent noise, the \MethodName{} actor is directly \emph{conditioned on the VLA's sampled reference action chunk and regularized toward it}, turning online RL into local refinement of a good VLA prior behavior policy rather than unconstrained search or implicit modulation of a diffusion process.
Together, these choices enable sample-efficient online RL on real robots---improving both success rate and execution speed within a few hours of practice.

\section{Preliminaries}
\label{sec:preliminaries}

\textbf{Vision-language-action models.}
Large-scale VLA models learn manipulation behaviors from diverse human demonstration datasets spanning tens of thousands of hours, in some cases augmented with non-robot vision-language data~\citep{pi05,geminirobotics,kim2024openvla}. A typical VLA comprises two components: (i)~a \emph{VLM backbone}, i.e., a vision-language model that encodes multi-modal inputs (images, language instructions, and proprioceptive state) into a shared token sequence, and (ii)~an \emph{action expert}, i.e., a diffusion-based module that attends to the backbone tokens and generates continuous actions through iterative denoising. We build on the $\pi_{0.6}$ model~\cite{pi06_model_card}. Given up to four camera images, a language instruction $\ell$, and proprioceptive state $\mathbf{s}^{\text{p}}_t$, $\pi_{0.6}$ produces an action sequence (referred to as an \emph{action chunk}): $\tilde{\mathbf{a}}_{t:t+H-1} = (\tilde{\mathbf{a}}_t, \ldots, \tilde{\mathbf{a}}_{t+H-1}) \in \mathbb{R}^{H \times d}$, a sequence of $H=50$ actions corresponding to 1\,s of control. We write $\pi_{\text{vla}}$ for the chunked policy produced by the pretrained VLA. In practice, the robot executes only a prefix of this chunk open loop (e.g., the first 20 steps) before re-planning from a new observation.
Due to the difficulty of some tasks (e.g., high-precision tasks) it can be challenging to collect large amounts of high-quality imitation learning data at scale for them, limiting the VLA's performance on these tasks. This motivates the online RL refinement method we develop in the next section. 

\textbf{Reinforcement learning and actor-critic methods.}
We formulate robot control as a Markov decision process (MDP) $(\mathcal{S}, \mathcal{A}, p, r, \gamma)$, where $\mathcal{S}$ is the state observation space, $\mathcal{A}$ is a continuous action space, $p(\mathbf{s}_{t+1} \mid \mathbf{s}_t, \mathbf{a}_t)$ denotes the transition dynamics, $r(\mathbf{s}_t, \mathbf{a}_t)$ is the reward function, and $\gamma \in [0,1)$ is the discount factor. The goal of RL is to learn a policy $\pi(\mathbf{a}_t \mid \mathbf{s}_t)$ that maximizes the expected discounted return:
$J(\pi) = \mathbb{E}_{\tau \sim \rho_\pi}\left[\sum_{t=0}^{T} \gamma^t r_t\right]$,
where $\rho_\pi(\tau)$ denotes the trajectory distribution induced by policy $\pi$. We assume access only to a \emph{sparse binary reward}: a human supervisor labels the end of each episode as success or failure, and we set $r_T = 1$ for success and $r_T = 0$ otherwise. The action-value function of a policy $\pi$ is
$
Q^\pi(\mathbf{s}_t, \mathbf{a}_t)=
\mathbb{E}_{\tau \sim \rho_\pi}
\left[
\sum_{t'=t}^{T} \gamma^{t'-t} r_{t'}
\;\middle|\;
\mathbf{s}_t, \mathbf{a}_t
\right].
$

In our setting, both policies and critics operate over action chunks
$
\mathbf{a}_{t:t+C-1} = (\mathbf{a}_t, \dots, \mathbf{a}_{t+C-1}) \in \mathbb{R}^{C \times d},
$
where $C$ denotes the RL chunk length (and $H$ denotes the chunk horizon predicted by the VLA). We choose $C < H$ to give the policy the ability to be more reactive. We define the chunked policy as $\pi(\mathbf{a}_{t:t+C-1} \mid \mathbf{s}_t)$ together with the corresponding chunk-level C-step estimate of the value
$
Q^\pi(\mathbf{s}_{t}, \mathbf{a}_{t:t+C-1}) = \sum_{t'=t}^{t+C-1} \gamma^{t'-t} r_{t'} + \gamma^C \mathbb{E}_{\mathbf{a}' \sim \pi | \mathbf{s}_{t+C}} \left[ Q^\pi(\mathbf{s}_{t+C}, \mathbf{a}') \right].
$
We build on classic off-policy actor-critic methods~\citep{heesssvg,haarnoja2018soft,fujimoto2018td3} that jointly train a stochastic actor $\pi_\theta$ and a critic $Q_\psi$. Crucially, learning is off-policy, using transitions stored in a replay buffer $\mathcal{B}$, regardless of the policy that generated them. This property is essential in our setting, where $\mathcal{B}$ aggregates data from the VLA policy, the RL learner, and human teleoperated interventions.
\section{\FullMethodName{}}
\label{sec:method}

Figure~\ref{fig:overview} summarizes our recipe for enabling fast and stable online RL from pretrained VLA models with \MethodName{}. The core idea is to maximally utilize the pretrained VLA to improve the efficiency of the RL training process. Training the entire VLA with online RL could be too compute- and sample-inefficient to produce an improved policy in just a few hours. Instead, we use the frozen VLA to provide the RL state representation, to supply reference actions and to guide exploration toward actions close to its own predictions, while still using a small actor and critic network.
We first adapt the VLA on a small amount of task-specific demonstration data, both to improve its initial task policy and to expose an RL token for downstream RL.
We then freeze the VLA and train lightweight off-policy actor and critic networks online, conditioning on both the RL token representation and the VLA's reference actions,
and regularize the learned policy to stay close to the VLA model.
Our method turns online RL into local refinement of promising behaviors rather than unconstrained search. This design gives the online RL method the efficiency of a small actor-critic algorithm, while retaining the representation and behaviors of the pretrained VLA model.


\subsection{Adapting the VLA to expose an RL interface}
\label{sec:rl_head}

\begin{figure}
\includegraphics[width=0.5\textwidth]{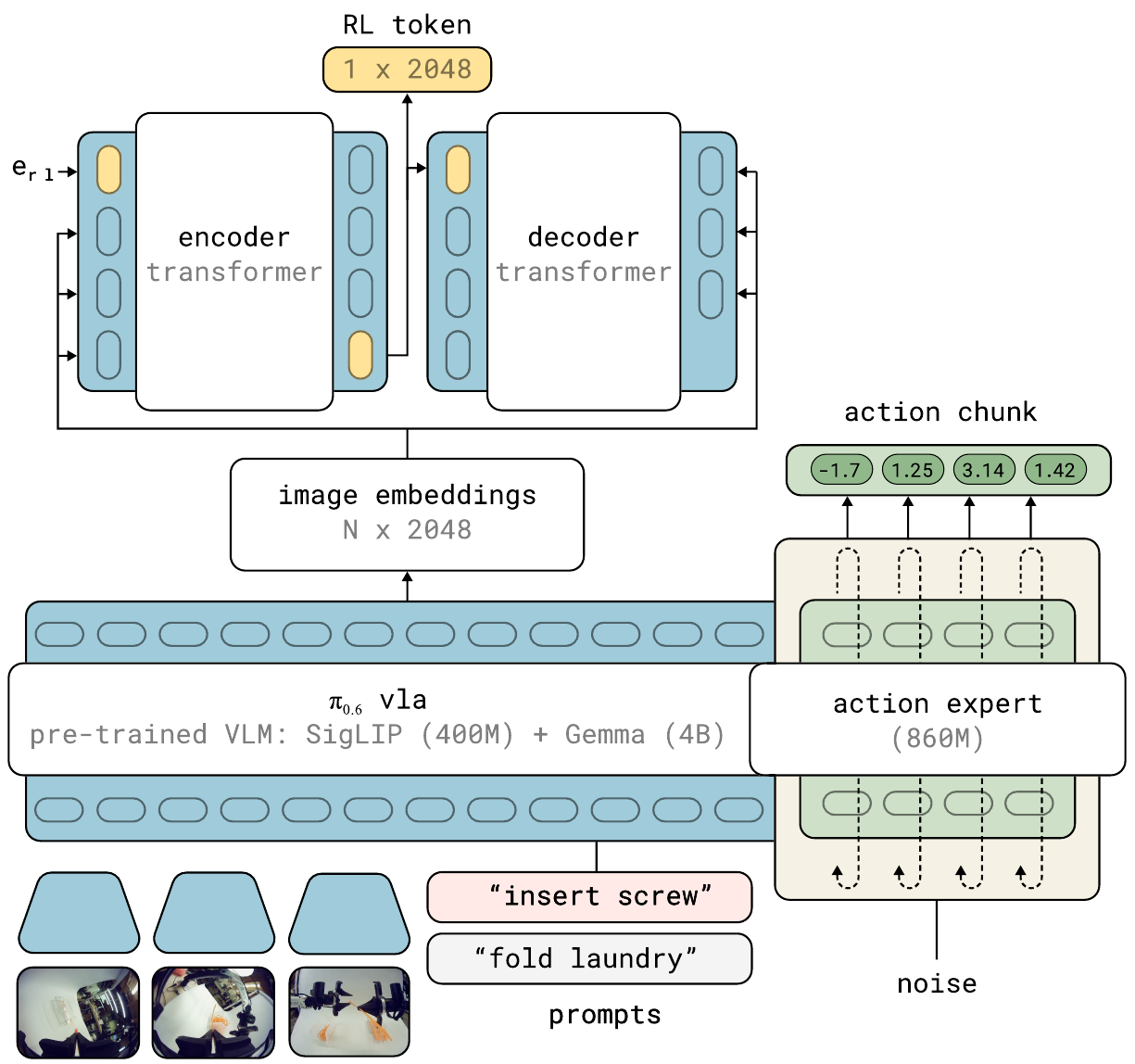}
    \captionof{figure}{\textbf{Details on RL token extraction.} \MethodName{} adds an encoder-decoder transformer to a pretrained VLA. It produces a compressed embedding of the VLA representation (the RL token). This representation then enables data and parameter efficient fine-tuning during online RL.}
    \label{fig:token}
\end{figure}

Sample-efficient online RL depends critically on the choice of state representation. Applying RL directly to the full VLA model is a poor match for rapid real-world adaptation: the representation is high-dimensional, and online updates to the billion-parameter model are both computationally expensive and sample-inefficient. At the same time, we would like to utilize the representation already contained inside a VLA after pretraining, since it was trained on large-scale web and robotic data and already contains information useful for generating actions for many tasks. However, it is generally not obvious which features from a transformer-based VLA make up a good representation for online RL and the embeddings in each transformer layer are high-dimensional. Our goal therefore is to compress the VLA representation into a compact embedding for RL that preserves task-relevant  information while remaining small enough for lightweight online actor-critic learning.

We achieve this by adding an \emph{RL token} (Fig.~\ref{fig:token}): a learned readout embedding that summarizes the VLA's knowledge into a small vector serving as the RL state.
Concretely, we obtain the RL token from a small additional transformer that we add to the pretrained VLA. We train this transformer in an encoder-decoder \citep{sutskever2014sequence} fashion, with the last input to the encoder being the RL token. Because the representation for the RL token must retain enough information to enable the decoder to reconstruct the inputs, it acts as a bottleneck.
Let $\mathbf{z} = f(s,\ell;\theta_{\text{vla}})$ denote the final-layer token embeddings produced by the pretrained VLA for state $s$ and language instruction $\ell$. The embeddings $\mathbf{z}$ decompose into $\mathbf{z}_{1:M} = \{\mathbf{z}_1,\dots,\mathbf{z}_M\}$, where each $\mathbf{z}_i$ corresponds to the embedding for one input token. 
We append a learned embedding $\mathbf{e}_\texttt{rl} = \mathbf{e}_\phi(\texttt{<rl>})$ to the sequence and process the augmented sequence with a lightweight encoder transformer $g_\phi$. The encoder output at the special-token position, denoted as $\mathbf{z}_{\text{rl}}$, is our RL token\footnote{In our experiments each task has a fixed language instruction, so we drop language embeddings in this step; the construction applies to all VLA embeddings in general.}
\begin{equation}
  \mathbf{z}_{\text{rl}} \;=\; g_\phi\!\bigl([\mathbf{z}_{1:M},\;\mathbf{e}_\texttt{rl}]\bigr)_{M+1}\,.
  \label{eq:readout}
\end{equation}
A decoder transformer~$d_\phi$ with a linear output projection $h_\phi$ is then trained to autoregressively reconstruct the original embeddings from $\mathbf{z}_{\text{rl}}$. Let $\bar{\mathbf{z}}_i = \mathrm{sg}(\mathbf{z}_i)$ denote the stop-gradient operation applied to VLA embeddings, then the autoregressive reconstruction objective on the demonstrations~$\mathcal{D}$ is given as:
\begin{equation}
  \mathcal{L}_{\text{ro}} = \mathbb{E}_{\mathcal{D}}\!\Bigl[\,
    \sum_{i=1}^{M}
      \bigl\lVert h_\phi\bigl(d_\phi([\mathbf{z}_{\text{rl}},\,\bar{\mathbf{z}}_{1:i-1}])\bigr)_{\!i}
        - \bar{\mathbf{z}}_i \bigr\rVert^2
  \,\Bigr].
  \label{eq:recon}
\end{equation}

We train the parameters~$\phi$ on a small task-specific demonstration dataset with the VLA considered frozen with respect to $\mathcal{L}_{\text{ro}}$, and (optionally) combine it with supervised fine-tuning of the VLA $(\theta_{\text{vla}})$. Afterwards, both $\theta_{\text{vla}}$ and $\phi$ are frozen, and online RL operates on the RL token representation $\mathbf{z}_{\text{rl}}$.


\subsection{Online RL to refine VLA action chunks}
\label{sec:chunk_rl}

After the initial adaptation stage we freeze both the VLA and the RL token representation. We then train lightweight actor ($\pi_\theta$) and critic ($Q_{\psi}$) networks online. Their inputs $x$ combine the RL token with any additional information useful to enable closed-loop control (e.g. the proprioceptive state of the robot). The critic model estimates the value of the state and action: $Q_{\psi}(\mathbf{x},\mathbf{a}_{1:C})\in \mathbb{R}$. Notably, rather than generating actions from scratch, the RL actor $\pi_\theta(\cdot | \mathbf{x}, \tilde{\mathbf{a}}_{1:C})$ is trained to refine action sequences $\tilde{\mathbf{a}}_{1:C}$ (referred to as action chunks) proposed by the VLA. 

\textbf{Training the critic.}  Our critic $Q_\psi( \mathbf{x},\mathbf{a}_{1:C})$ takes as input the state and the action chunk $\mathbf{a}_{1:C}$. We train the critic with standard off-policy temporal-difference learning on action chunk transitions sampled from the replay buffer $\mathcal{B}$:
\begin{equation}
\begin{aligned}
\mathcal{L}_Q&=\mathbb{E}_{(\mathbf{x}, \mathbf{a}_{1:C}, \mathbf{x}') \sim \mathcal{B}}
\Big[\big(\hat{Q}-Q_\psi(\mathbf{x}, \mathbf{a}_{1:C})
\big)^2\Big], \\
\hat{Q}&=\sum_{t'=1}^{C} \gamma^{t'-1} r_{t'}+\gamma^C\mathbb{E}_{\mathbf{a}' \sim \pi_\theta}
\Big[Q_{\psi'}(\mathbf{x}', \mathbf{a}')\Big].
\end{aligned}
\label{eq:critic_loss}
\end{equation}
where the input state is $\mathbf{x} =(\mathbf{z}_\text{rl}, \mathbf{s}^\text{p})$, and $\mathbf{s}^\text{p}$ denotes the proprioceptive state information, $\mathbf{z}_\text{rl}(\mathbf{s})$ denotes the RL token extracted for state $\mathbf{s}$; $\mathbf{x}'$ denotes the next input state; $\mathbf{a}' \sim \pi_\theta$ denotes taking a sample from the RL policy.  In practice, we follow TD3 \citep{fujimoto2018td3} and $\psi'$ are the parameters of the target network.

\textbf{Training the RL Policy.}
Our actor network $\pi_\theta(\cdot| \mathbf{x},\tilde{\mathbf{a}}_{1:C})$ produces a Gaussian action distribution over action chunks. It takes in the input state  \emph{and} a reference action chunk $\tilde{\mathbf{a}}_{1:C}$, and produces the action distribution: 
\begin{equation}
\pi_\theta\bigl(\mathbf{a}_{1:C} \mid \mathbf{x}, \tilde{\mathbf{a}}_{1:C}\bigr)
  =
  \mathcal{N}\Bigl(
    \mu_\theta\bigl(\mathbf{x}, \tilde{\mathbf{a}}_{1:C} \bigr),
    \sigma^2 \mathbf{I}
  \Bigr),
  \label{eq:policy}
\end{equation}
where, as before, $\mathbf{x} =(\mathbf{z}_\text{rl}, \mathbf{s}^\text{p})$.
Conditioning on $\tilde{\mathbf{a}}$ exposes the actor directly to the VLA’s predicted actions, so that online RL refines a strong initial proposal rather than learning from scratch. 
A second benefit is that the sampled reference chunk preserves mode information from the VLA’s multimodal action distribution, which would otherwise be difficult for a unimodal Gaussian actor to recover~\cite{fql_park2025}.
We further stabilize learning by regularizing its action toward the reference action. Concretely, we optimize the actor to maximize critic value while staying close to the VLA reference chunk $\tilde{\mathbf{a}}$, similar in spirit to KL-regularized RL methods (see e.g.~\citep{peng2020learning,PetersREPS,Abdolmaleki2018MaximumAP,Dayan1997UsingEM,levine2018reinforcementlearningcontrolprobabilistic}). This effectively turns online RL into local action editing around the VLA generated action distribution, rather than an unconstrained search over high-dimensional action chunks. The objective for learning the RL policy is given as
\begin{equation}
\begin{aligned}
\mathcal{L}_{\pi}(\theta)
&=
\mathbb{E}_{\substack{\mathbf{s} \sim \mathcal{B} \\
\mathbf{a}_{1:C} \sim \pi_\theta}}
\left[
-\,Q_\psi(\mathbf{x}, \mathbf{a}_{1:C})
+\beta \left\|\mathbf{a}_{1:C}-\tilde{\mathbf{a}}_{1:C}\right\|_2^2
\right], \\
\qquad
& \qquad \tilde{\mathbf{a}}_{1:C}\sim\pi_{\mathrm{vla}}(\cdot\mid\mathbf{s}, \ell),
\end{aligned}
\label{eq:actor_loss}
\end{equation}
where the coefficient $\beta$ controls how strongly the actor is regularized towards the sampled VLA action. 

\textbf{Reference action dropout.} A practical failure mode of reference-action conditioning is that the actor may simply copy $\tilde{\mathbf{a}}$ instead of learning to improve it. This is especially likely before the critic becomes informative, since both conditioning on $\tilde{\mathbf{a}}$ and regularizing towards it encourage the actor to remain close to the VLA proposal. To prevent this, we apply \emph{reference action dropout}: for a random subset of transitions in each training batch, we replace the reference chunk with zeros before passing it to the actor. This forces the actor to maintain an independent action-generation pathway, while still allowing it to exploit the VLA action distribution whenever the reference chunk is present. In practice, once the critic provides useful signal, the actor naturally learns to deviate from the reference whenever doing so increases predicted value.

\begin{algorithm}[t]
\caption{\MethodName{}}
\label{alg:main}
\begin{algorithmic}[1]
\REQUIRE \textcolor{gray}{\textit{Frozen VLA backbone $f_{\theta_\text{vla}}$ and VLA action distribution $\pi_{\text{vla}}$;
demo data $\mathcal{D}$, chunk length $C$, replay buffer $\mathcal{B}$,
warmup steps $N_{\text{warm}}$, ratio $G$, VLA fine-tuning weight $\alpha$, policy constraint $\beta$.}} 
\vspace{2pt}

\STATE \textbf{Train RL token and (optionally) fine-tune the VLA}
\STATE Train $\phi$ using $\mathbf{z}_i=f_i(\mathbf{s}, \ell, \theta_\text{vla})$, $\mathbf{z}_{\text{rl}} = g_\phi([\mathbf{z}_{1:M}, \mathbf{e}_\text{rl}])_{M+1}$, and $\theta_\text{vla}$ (only if $\alpha > 0$).
\[
\mathcal{L}_{\text{ro}}(\phi) = \mathbb{E}_{\mathcal{D}}\!\Bigl[\,
    \sum_{i=1}^{M}
      \bigl\lVert h_\phi\bigl(d_\phi([\mathbf{z}_{\text{rl}},\,\bar{\mathbf{z}}_{1:i-1}])\bigr)_{\!i}
        - \bar{\mathbf{z}}_i \bigr\rVert^2
  \,\Bigr].
\]
\STATE 
\[
\phi, \theta_{\text{vla}} = \arg \min_{\phi, \theta_{\text{vla}}} \mathcal{L}_{\text{ro}}(\phi) + \alpha \mathcal{L}_{\text{vla}}(\theta_\text{vla})
\]
\vspace{2pt}

\STATE \textbf{Train RL actor and critic}
\STATE Initialize critic $Q_\psi$ and RL Policy $\pi_\theta$.
\FOR{environment steps $t=0,C,2C\dots$ }
  \STATE Sample VLA reference chunk $\tilde{\mathbf{a}}_{t:t+C-1}\sim \pi_{\text{vla}}(\mathbf{s}_t)$.
  \STATE Form RL state $\mathbf{x}_t = (\mathbf{z}_{\text{rl}}(\mathbf{s}_t), \mathbf{s}^p_t)$.
  \STATE $
\mathbf{a}_{t:t+C-1} \leftarrow
\begin{cases}
\mathbf{a}^{\text{human}} & \text{if intervention} \\
\tilde{\mathbf{a}}_{t:t+C-1}        & \text{if } t < N_{\text{warm}} \\
\sim \pi_\theta(\,\cdot \mid \mathbf{x}_t,\tilde{\mathbf{a}}\,) & \text{otherwise}
\end{cases}
$
  \STATE Execute $\mathbf{a}_{t:t+C-1}$ and observe $r_t$, $\mathbf{s}_{t+1}$, $\mathbf{s}^p_{t+1}$
\STATE $
\tilde{\mathbf{a}}_{t:t+C-1} \leftarrow \mathbf{a}^{\text{human}} \quad \text{if intervention} $
  \STATE Store transition in $\mathcal{B}$:\quad $\langle\mathbf{x}_t,\mathbf{a}_{t:t+C-1},\tilde{\mathbf{a}},r_t,\mathbf{x}_{t+1}\rangle$

  \FOR{$g=1,\dots,G$}
    \STATE Sample batch of data $\mathrm{b} \sim \mathcal{B}$.
    \STATE Compute target Q values
    \[
    \hat{Q} = \sum_{t'=1}^C \gamma^{t' - 1} r_{t'} + \gamma^{C} \mathbb{E}_{\mathbf{a}' \sim \pi_\theta} \big[ Q_{\psi'}(\mathbf{x}', \mathbf{a}') \big]
    \]
    \STATE Train Critic with TD backup (Eq. \eqref{eq:critic_loss})
    \[
    \mathcal{L}_Q(\psi) = \mathbb{E}_{\mathrm{b}} \Big[ 
    \big( \hat{Q} -   Q_\psi(\mathbf{x}, \mathbf{a}) \big)^2 \Big]
    \]
    \STATE Train Policy $\mathbf{a} \sim \pi_\theta(\cdot\mid \mathbf{s},\tilde{\mathbf{a}})$ (Eq. \eqref{eq:actor_loss})
    \[
      \mathcal{L}_{\pi}(\theta)=\mathbb{E}_\mathrm{b}\Bigl[
      -Q_\psi(\mathbf{x},\mathbf{a})
      +\beta\|\mathbf{a}-\tilde{\mathbf{a}}\|_2^2
      \Bigr] 
    \]
  \ENDFOR
\ENDFOR
\end{algorithmic}
\end{algorithm}

\section{The Complete System}
\label{sec:system}

\begin{figure*}[!htbp]
    \centering
\includegraphics[width=1\linewidth]{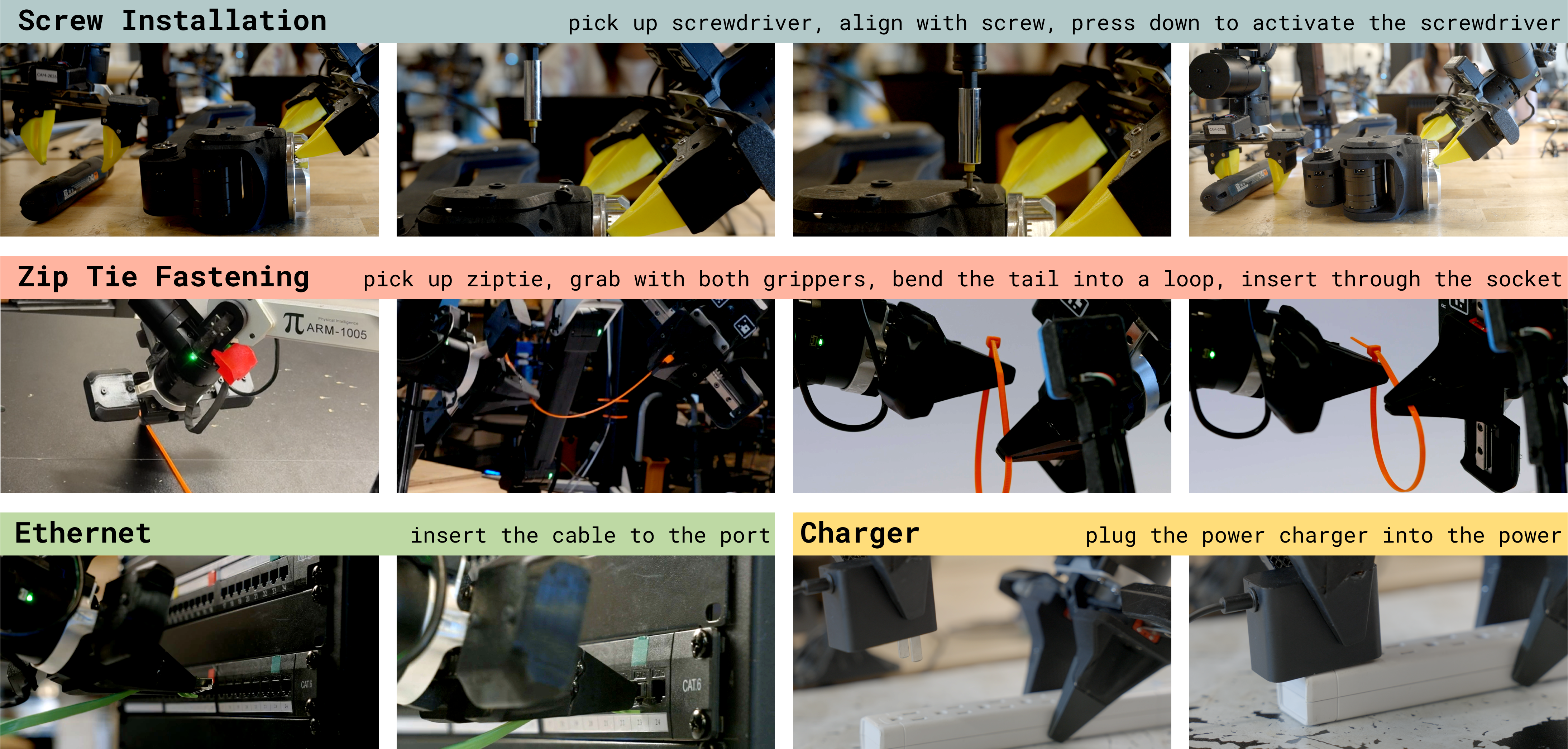}
    \caption{\textbf{The tasks in our experiments}: each task contains a critical phase that requires high precision: (top) using a screwdriver to install a screw, (middle) fastening a zip tie, (bottom) plugging in an Ethernet cable and plugging in a charger.}
    \label{fig:tasks}
\end{figure*}

Algorithm~\ref{alg:main} summarizes our full training loop. After an initial warmup stage to collect episodes with the base VLA policy, training alternates between collecting experience on the robot and performing off-policy actor-critic updates from replay. The replay buffer aggregates VLA warmup data, online RL rollouts, and optional human interventions. In addition, a human supervisor provides sparse success/failure labels. The steps are described in detail below. 

\textbf{Warmup.} After training the RL token representation (Sec.~\ref{sec:rl_head}), we pre-fill the replay buffer $\mathcal{B}$  by rolling out the VLA reference policy for $N_{\text{warm}}$ environment steps. This gives the critic an initial learning signal and ensures that online RL begins from competent VLA behavior.

\textbf{Rollout.} At each action chunk boundary during online collection, the frozen VLA produces a reference chunk $\tilde{\mathbf{a}}_{1:H}$ and the RL token module extracts $\mathbf{z}_{\text{rl}}$. The actor then outputs an action chunk $\mathbf{a}_{1:C} \sim \pi_\theta(\cdot \mid \mathbf{x}, \tilde{\mathbf{a}}_{1:C})$.
To accelerate learning of contact-rich or safety-critical behaviors, a human operator may optionally intervene by providing teleoperated commands $\mathbf{a}^{\text{h}}_{1:C}$ that overwrite the actor output for the duration of the intervention. When this occurs, the intervention replaces the VLA reference in the replay buffer. In all cases, each transition stored in $\mathcal{B}$ includes the executed action and the corresponding reference, enabling the actor to learn from both autonomous rollouts and human corrections.


\textbf{Subsampling Action Chunks.} While the RL policy uses an action chunk length of $C$, we obtain observations for each intermediate step. We can thus increase data and improve learning efficiency by storing intermediate steps into the replay buffer. Concretely we pick a stride of $2$ and save transitions corresponding to $<\mathbf{x_0}, \mathbf{a_{0:C}}>, <\mathbf{x_2}, \mathbf{a_{2:C+2}}>, <\mathbf{x_4}, \mathbf{a_{4:C+4}}>, \dots$ to the replay buffer. Note that, due to the off-policy nature of our RL algorithm we can use all action chunks (including VLA-generated actions and human interventions).

\textbf{Update.} Policy updates are performed off-policy from the replay buffer according to Algorithm~\ref{alg:main}. To remain compute and time-efficient during training, we perform the rollouts and learning asynchronously. In practice, we perform two critic updates for each actor update, and begin learning shortly after the warmup phase. We use a high update-to-data ratio of $5$, which is essential in the low-data online regime.

\textbf{Targeted improvement of critical phases.}
For practicality and efficiency of learning, we apply \MethodName{} to improve the critical phase of each task we consider -- corresponding to the most difficult sections which require high-precision -- and let the base VLA perform the easier parts of the task.
Concretely, each episode starts by executing the base model. During data collection, a human operator can then choose at which point to hand the control from the base VLA to the RL policy. This is analogous to human intervention decisions in interactive imitation learning~\citep{kelly2019hgdaggerinteractiveimitationlearning}. Our system then applies RL to the chosen task segment, storing and training on transitions during this critical phase, until receiving a terminal signal from the human operator that indicates success or failure of the RL task. This concentrates data collection and credit assignment on the part of the behavior where online adaptation matters most. 
To enable autonomous execution at test time, we can conclude training with a final short fine-tuning phase of the VLA where we ask it to additionally predict when to hand over execution to the RL policy (using the human interventions as labels). We can then automatically trigger the switch of policies at test-time.

\section{Real-World Experiments}
\label{sec:experiments}

We evaluate \MethodName{} on four real-world manipulation tasks that require dexterous control and sub-millimeter precision. A pretrained VLA provides a strong initialization for most parts of these tasks, but success and speed ultimately depend on refining the critical contact-rich phase that demands the most precision. Our experiments test whether our method can deliver such improvements under the practical constraints that motivate the method: limited robot interaction time, sparse human supervision, and lightweight online learning.

We structure the evaluation around the following questions:
\begin{enumerate}
    \item[\textbf{Q1.}] Can \MethodName{} improve manipulation performance over the base VLA model?
    \item[\textbf{Q2.}] How does \MethodName{} compare to alternative RL approaches on these tasks?
    \item[\textbf{Q3.}] How much does each component of the method---the RL token, chunked action prediction, policy regularization, and reference-action pass-through---contribute to the method's performance?
    \item[\textbf{Q4.}] Does \MethodName{} enable the policy to discover a better strategy and how does its strategy compare to the original demonstration data? 
\end{enumerate}
\subsection{Tasks and Setup}
\label{sec:tasks}

We evaluate our method on the following tasks (Fig.~\ref{fig:tasks}):

\begin{itemize}[leftmargin=*, itemsep=2pt]
    \item \textbf{Screw installation.} The robot must use an electric screwdriver to drive an M3 screw into a threaded receptacle. This requires sub-millimeter alignment between the screw head and screwdriver tip. The task is especially difficult because (1) the screw may not always sit perfectly upright, (2) when holding the screwdriver, any rotation of the end effector is amplified by the 10\,cm distance between the screwdriver tip and the grasp point, and (3) critical visual cues are visible primarily from the wide-angle wrist camera on the opposite arm, presenting a challenging perceptual problem.
    
    \item \textbf{Zip tie fastening.} The robot must thread a zip tie tail through its narrow locking slot. This task involves coordinated bimanual control of a deformable object with tight tolerances. Successful insertion requires inferring the position of the tip and slot solely from the wrist cameras and executing with millimeter precision.
    
    \item \textbf{Ethernet insertion.} The robot must insert an Ethernet connector into a recessed port. This requires accurate positional and angular alignment, followed by a firm and decisive insertion motion. Small orientation errors or hesitant contact typically cause the connector to catch on the housing rather than insert into the port, making success sensitive to both precision and contact dynamics.
    
    \item \textbf{Charger insertion.} The robot must align and insert a charger into a power strip. The task is difficult because the policy must achieve centimeter-level alignment while not always having clear observability of the prongs and socket. Small alignment errors often lead to repeated probing or failed insertion attempts.

\end{itemize}

Each task includes grasping, repositioning, and alignment, and spans 30--120\,s (at 50\,Hz roughly 1500--6000 control steps). For each task, we identify the \emph{critical phase}---the insertion, fastening, or rotation segment---where precision requirements are the highest and where the base VLA most frequently slows down or fails. These phases typically last 5--20\,s (250--1000 control steps). 

\textbf{Critical-phase evaluation.}\quad
Since our method is designed to improve exactly these critical phases, we first focus our evaluation on comparing methods and ablations only on the critical phase. In this setting, episodes begin after being reset to a partially completed task state right before the critical phase, using a slightly randomized set of initial configurations. For example, in zip tie fastening, the robot begins already holding the two ends of the zip tie before the insertion attempt begins. This setup isolates the precision-critical segment where RL is expected to matter most, and reduces confounding variance from earlier phases of the task, such as grasping and transport, which are already handled reasonably well by the base VLA. Each agent is evaluated over 50 episodes per task in this controlled setting.

\textbf{Full-task evaluation.}\quad
Controlled critical-phase evaluation is useful for isolating the bottleneck that our method is designed to improve, but it does not capture the full variability of long-horizon execution. We therefore additionally evaluate full-task performance in a more realistic setting, where the robot starts from its ``home position'', executes the earlier stages of the task with the base policy, and enters the critical phase under the state variation induced by that execution. This setting is substantially harder, since the RL-improved behavior must remain effective under the broader distribution of states produced by the preceding policy. For the full-task training, we start by having RL focus on the critical phase with small randomization, and then move on to the full-task setting.

\textbf{Experiment Details.}\quad
The RL policy inputs consist of the RL token (produced from two wrist camera images and one base-camera image), and additional proprioceptive state. Depending on the task, this auxiliary state may include joint position (screw), end-effector pose (zip tie, Ethernet and charger). We use $\pi_{0.6}$~\cite{pi06_model_card} as the base VLA policy. The robot runs at a control frequency of 50\,Hz. With a 14-dimensional per-timestep action space, this corresponds to a 140-dimensional chunked action for the RL actor. We provide more implementation details in App.~\ref{app:exp}. 



\subsection{Baselines and Ablations}
\label{sec:baselines}

\begin{figure*}
    \centering
    \includegraphics[width=1\linewidth]{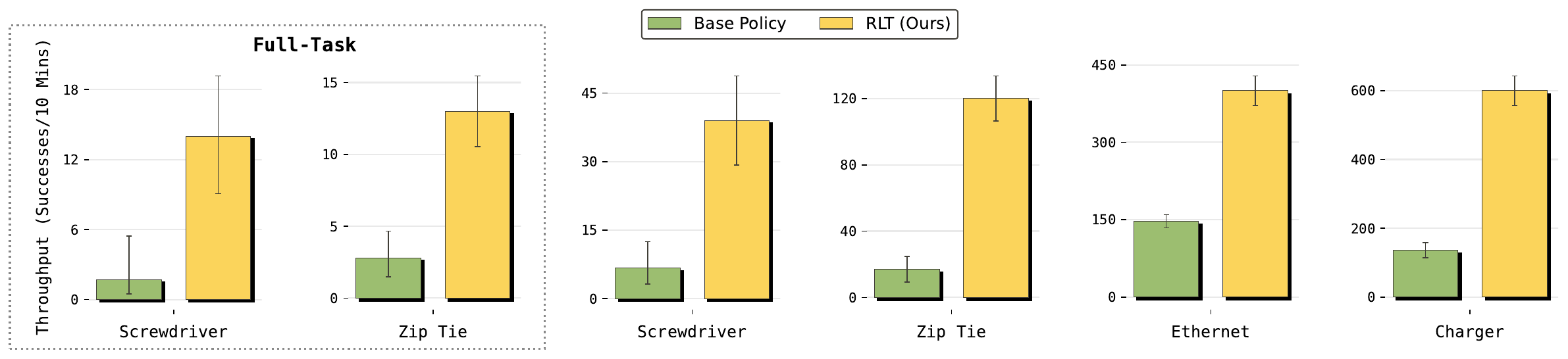}
    \caption{\textbf{\MethodName{} increases throughput significantly} over the base VLA policy, improving both the speed and consistency of the critical phase of each task. The improvement is especially pronounced for the harder tasks where the VLA policy is prone to making mistakes.}
    \label{fig:results_throughput}
\end{figure*}

\begin{figure*}
    \centering
    \includegraphics[width=1\linewidth]{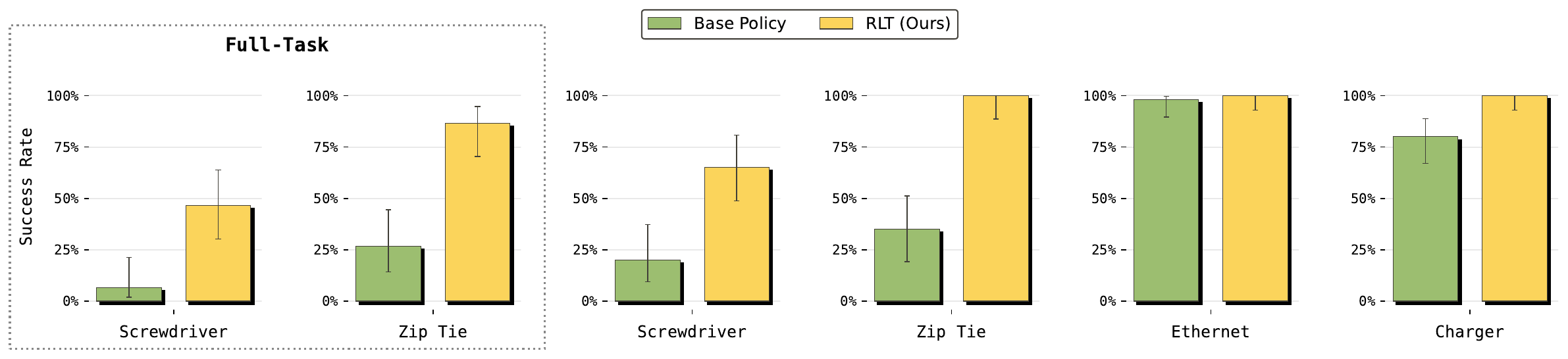}
\caption{\textbf{\MethodName{} can boost success rates across multiple tasks}. Where the VLA is already competent (e.g., the Ethernet task) it maintains success rate and increases throughput. For tasks that are challenging for the base VLA policy (screwdriver and zip tie) \MethodName{} leads to a significant improvement in success.}
    \label{fig:results_success}
\end{figure*}

We start from a pretrained VLA model $\pi_{0.6}$~\cite{pi06_model_card}.
For each task, we collect $1$--$10$ hours of teleoperated demonstrations. We then fine-tune the VLA model while training our RL token representation.
This produces the base VLA policy that we use throughout all of the experiments. We run RL training for between 400 and 1000 episodes depending on the task difficulty. Excluding reset and various overhead, each of the experiments produces approximately 15 minutes to 5 hours of actual robot data. We measure the performance in terms of the success rate on each task, judged by a binary reward signal from a human operator. We also report throughput, the number of successful task completions per 10 minute interval, in order to evaluate improvements in terms of both robustness and speed. We evaluate all tasks on their critical phases, and we evaluate the two harder tasks---the screw and zip tie tasks---in the full-task setting.

We compare \MethodName{} to four baseline methods that improve policies from experience. For fair comparisons, we train each RL method with the same amount of data (see App.~\ref{app:baseline}).

\begin{itemize}[leftmargin=*, itemsep=2pt]
    
    \item \textbf{HIL-SERL}~\cite{luo2024precise}: Like our method, HIL-SERL trains a small actor and critic with a combination of experience and interventions, but unlike \MethodName{}, it does not use representations from a pretrained VLA, instead using a simple ResNet encoder pretrained for standard computer vision tasks.
    \item \textbf{Probe-Learn-Distill}~\cite{xiao2025selfimprovingvisionlanguageactionmodelsdata}: PLD learns a residual policy that outputs a residual for each single-step action. It scales the residual by a hyperparameter and sums it with one step from the frozen VLA's action prediction to execute. 
    \item \textbf{DSRL}~\cite{Wagenmaker2025DSRL}: DSRL learns an online RL policy in the latent noise space of a flow VLA model. It ``steers'' the VLA action generation by selecting the noise to feed into the frozen VLA model's action generator. This method implicitly constrains exploration to those actions that can be generated by the VLA, and explores among its modes.
    \item \textbf{DAgger}~\cite{rossdagger,kelly2019hgdaggerinteractiveimitationlearning}: We fine-tune the base VLA model on the human intervention data collected during our training. 
\end{itemize}
We also isolate the contribution of each component of our method by removing them individually:
\begin{itemize}[leftmargin=*, itemsep=2pt]
    \item \textbf{w/o RL token}: Replace the RL token with a frozen, ImageNet pretrained ResNet-10 encoder from \citep{luo2024serl}.
    \item \textbf{w/o Chunk}: The RL policy outputs single-step actions ($C{=}1$) instead of action chunks. Because this policy needs to run at 50\,Hz and querying the base VLA model at 50\,Hz would be infeasible, we have to replace the RL token with the ResNet-10 encoder. 
    \item \textbf{w/o BC Regularizer}: Set $\beta{=}0$ in Eq.~\eqref{eq:actor_loss}; the policy is trained with the $Q$-function alone.
    \item \textbf{w/o Pass-Through}: Remove $\tilde{\mathbf{a}}$ from the policy input in Eq.~\eqref{eq:policy}; the RL actor generates actions from state and RL token alone.
\end{itemize}

\subsection{Experimental Results}
\label{sec:results}


\begin{figure}
    \centering
    \includegraphics[width=\linewidth]{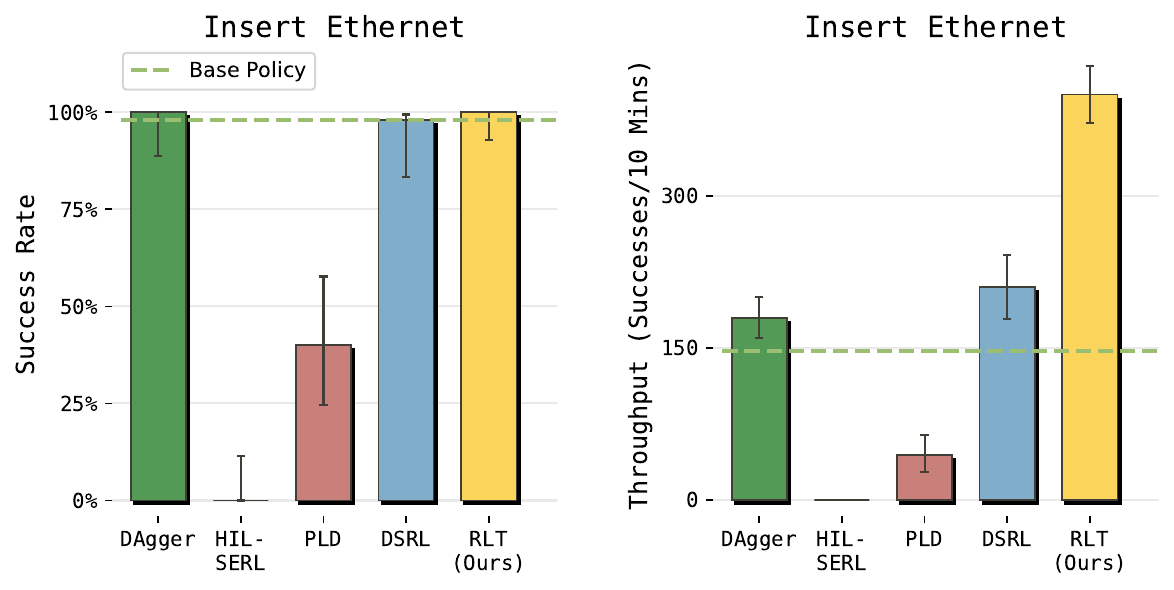}
    \caption{\textbf{Comparison to other RL algorithms.} We compare \MethodName{} against several baselines from the recent RL literature. Methods that consider only single actions, rather than action chunks, (HIL-SERL, PLD) perform poorly. DSRL leads to high success but significantly lags behind in throughput.}
    \label{fig:baseline_comparison}
\end{figure}

\textbf{Q1: Does online RL improve over the base VLA policy?}
We evaluate our method in two regimes: the \emph{controlled} setting that isolates the critical phase and the \emph{full-task} setting that requires the RL policy to be more robust.
\emph{Online RL improves success rate and execution speed of the base model} in both settings. In the controlled setting, \MethodName{} consistently improves the critical phase of all four tasks. Even on the relatively easier charger and Ethernet tasks, where the base policy already achieves good reliability, the policy learned by \MethodName{} is about $3{\times}$ faster in the critical phase. The boost to success rate is more pronounced on the harder zip tie and screwdriver tasks.  In the full-task evaluation, overall success rates are lower due to compounding errors from earlier parts of the task (grasping/lifting objects etc.), but \MethodName{} still improves the success rate by 40\% on the screwdriver task and 60\% on the zip tie task.

\textbf{Q2: How does \MethodName{} compare to alternative methods?}
As shown in Fig.~\ref{fig:baseline_comparison}, \emph{\MethodName{} results in a significant boost to throughput compared to baselines}. We compare against four baselines on the Ethernet task. HIL-SERL and PLD---both single-step online RL methods---fail to learn effectively on this task, which spans hundreds of steps with a sparse reward. Without action chunking, the horizon of the task is very long, and the value function updates are ineffective at propagating the sparse reward signal. For this simpler task, DAgger and DSRL achieve success rates comparable to \MethodName{} (Fig.~\ref{fig:baseline_comparison}), but provide significantly less improvement in terms of speed. DAgger is an imitation learning method, and is limited to the speed of the human demonstrations and interventions. DSRL is an RL method that strongly constrains the policy to remain close to the base VLA, providing stable training but comparatively less potential for improvement.  In contrast, \MethodName{} matches the base policy's high success rate while reducing mean steps to completion by $2{\times}$ over the base policy.

\begin{figure}
    \centering
    \includegraphics[width=1\linewidth]{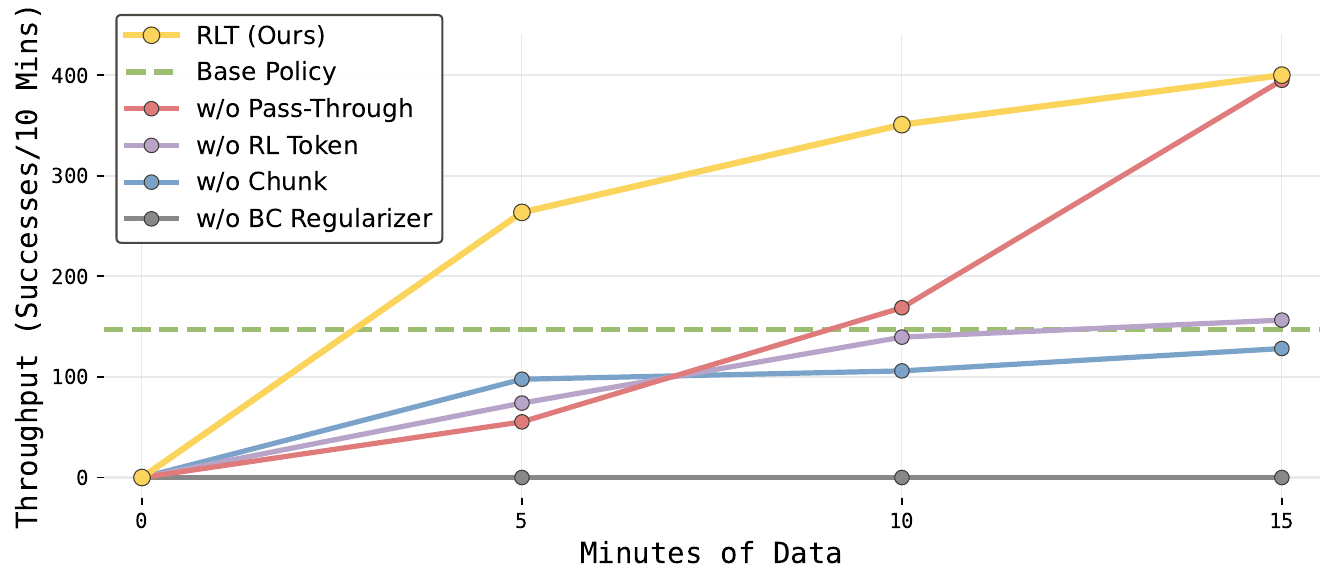}
    \caption{\textbf{Throughput at different points in training for Ethernet task.} The ablation study shows that each part of our method is important for good performance, and the full system learns fastest and performs best at the end. Notably, \MethodName{} outperforms the alternative policy after consuming only 5 minutes of data on the critical part of the task (total experiment time $\sim$ 40mins). Dropping the reference action from the actor input (``w/o Pass-Through'') still allows reaching the best final performance, but at the cost of slower learning and significantly more failures over the course of training. }
    \label{fig:ablation_throughput}
\end{figure}

\textbf{Q3: How much does each component contribute?}
\emph{All four design choices---RL token, action chunks, the BC regularizer, and reference-action pass-through---contribute meaningfully}.
We validate that each component of our recipe provides a positive contribution (Fig.~\ref{fig:ablation_throughput}): replacing the RL token with a ResNet-10 encoder reduces throughput by $50$\%, confirming that our token encodes manipulation-relevant structure that an off-the-shelf encoder trained on standard computer vision tasks does not provide. Replacing chunks ($C{=}10$) with single-step actions dramatically increases the effective horizon of the task, since the value function needs to perform credit assignment over a much longer horizon. It also would make it infeasible to run our method with the RL token. In practice, the single-step variant cannot reliably match the base policy performance. 
Removing the BC regularizer ($\beta{=}0$) causes the largest single drop in performance, since it forces the actor to explore the full action space with only gradients from the $Q$-function. Removing the reference-action pass-through slows down learning, leads to early exploration drift, and occasionally degenerate behaviors. This ablation does eventually match the performance of \MethodName{} for this simpler task, but experiences more failures during the training process, as seen on the learning curve in Fig.~\ref{fig:ablation_throughput}.


\begin{figure}
    \centering
\includegraphics[width=1\linewidth]{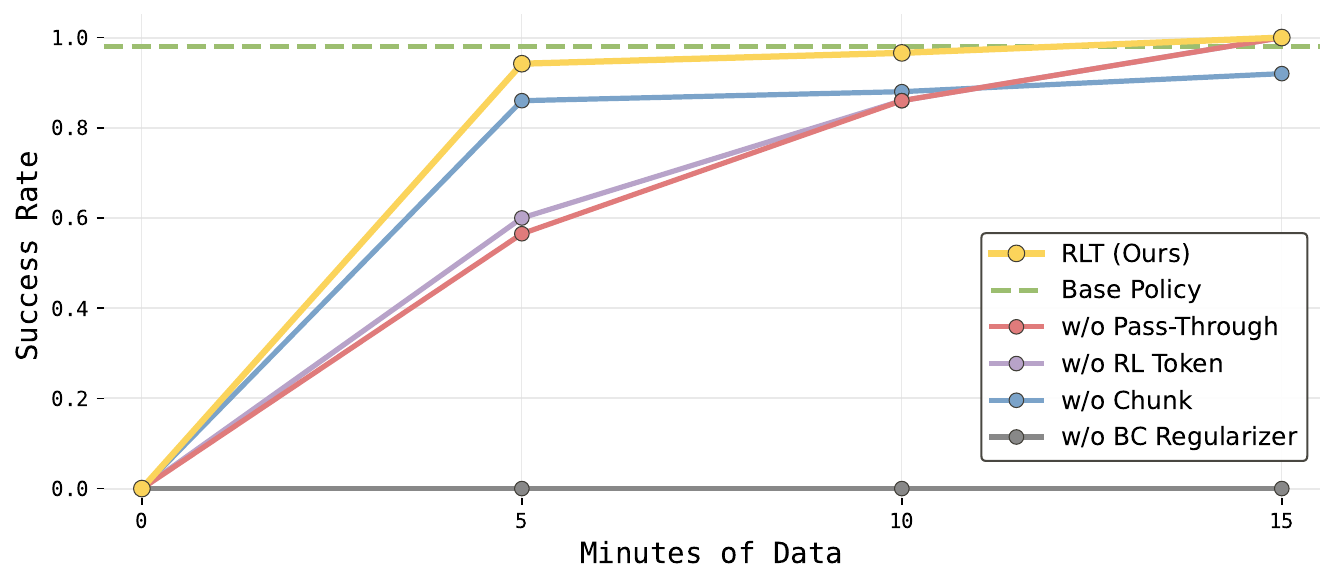}
    \caption{\textbf{Success rate evaluation during training for the Ethernet task.} \MethodName{} quickly matches the success rate of the VLA policy on the Ethernet insertion task, while boosting throughput. Not using the reference-action pass-through or not using the RL token leads to slower learning.}
    \label{fig:placeholder}
\end{figure}

\textbf{Q4: Does \MethodName{} lead to more effective emergent strategies?}
\begin{figure}
    \centering
    \includegraphics[width=1\linewidth]{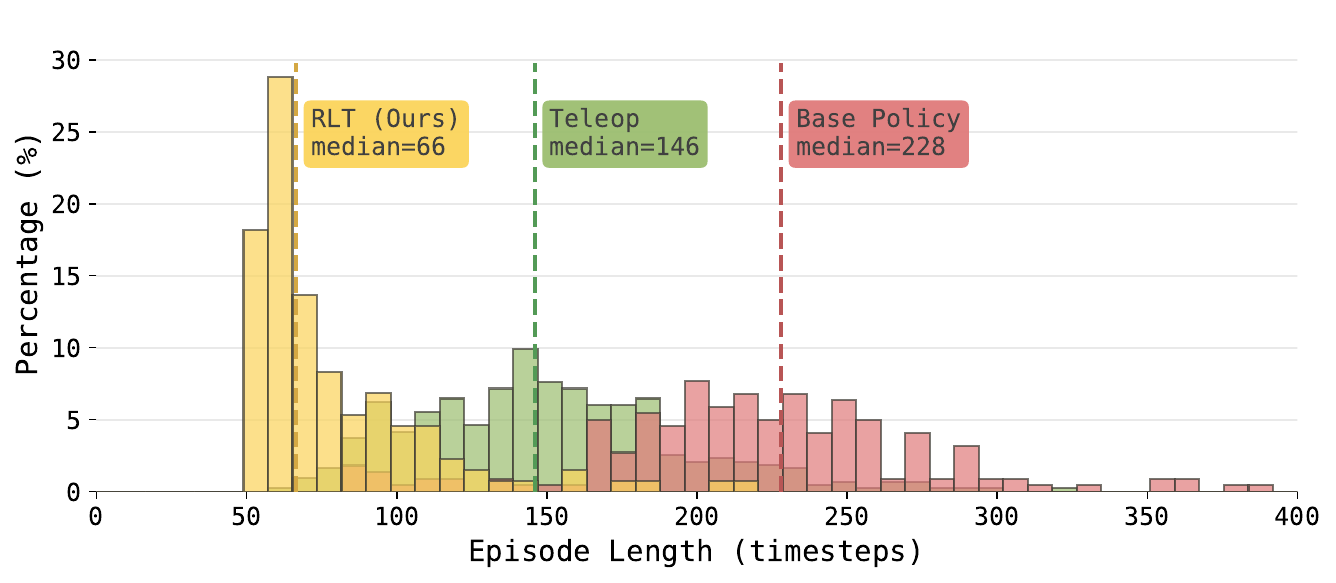}
    \caption{\textbf{Speed on the Ethernet task.} \MethodName{} significantly improves the speed of the Ethernet task. The final policy is faster even than the demonstrations produced by expert teleoperators, and significantly faster than the base VLA model. Half of the RL episodes during the critical insertion phase (yellow) are faster than all of the teleoperated demonstrations (green).}
    \label{fig:speed}
\end{figure}
Beyond aggregate metrics, the  effect of online RL is a qualitative change in \emph{how} the robot performs the task. On the critical phase for the Ethernet task, we visualize the distribution of speeds for the teleoperated demonstrations, the base policy, and the final RL policy (Fig.~\ref{fig:speed}). The base VLA frequently exhibits ``probing'' behavior near contact: it approaches the target, retreats slightly, readjusts, and attempts again---sometimes cycling through several such attempts before succeeding. \MethodName{} instead approaches the port, and inserts the connector in a fluid motion. Even when it fails on the first attempt, \MethodName{} applies  pressure and wiggles the connector slightly to exploit compliance, leading to faster insertion. This behavior is not seen in the demonstration data and emerges purely from online exploration, illustrating that the method can go beyond imitating human strategies.

\section{Conclusion}
\label{sec:conclusion}

We presented \MethodName{}, a method for fast online RL on top of representations extracted from a large pretrained VLA. By training the VLA to expose a compact representation, our method enables a lightweight actor and critic to improve highly precise and delicate tasks with just a few hours of real-world practice. Across four difficult tasks that require precision and speed, \MethodName{} consistently improves both the success rate and execution speed, achieving up to 3× speedup on the hardest phase of each task and, in some cases, surpassing expert human teleoperation speed through strategies that emerge from online RL.


While \MethodName{} provides fast and efficient learning, it does require additional human intervention during training to provide reward signals, intervention corrections, and switching between RL (for the critical phase) and the base policy (for the other phases). In principle, some of these components can be automated, for example by using reward models and progress prediction. Developing a fully autonomous RL improvement pipeline based on \MethodName{} is a promising direction for future work. More broadly, we believe that our method represents an important step toward robotic systems that not only learn from demonstration data, but can improve directly on the job. When improvement is fast and reliable, it is enough for the pretraining phase of the VLA to simply provide a good initialization for downstream exploration, while the most successful and performant strategy can be discovered through reinforcement learning. We hope that \MethodName{} will serve as a step toward this future.


\section*{Acknowledgments}

Robotics is a team effort. We are grateful to all of the people at Physical Intelligence who contributed to the many facets of this work, including hardware, data collection, robot operations, and robot infrastructure. We thank Liam Murphy and Cameron Myers for their help on the gripper design. We thank the robot operators and the operation and annotation team at PI. We thank Connor Jacobsen for help with the website and blog post, Brian Ichter for help with figures, Kyle Vedder for proofreading, Claudio Guglieri for help with visualizations for the blog post, and Donald Jewkes and Thomas Burton for help with filming and editing of the video.
%

\bibliographystyle{unsrtnat}
\bibliography{main}  


\newpage

\appendix

\subsection{Contributions}

CX, LK started the project. CX built the infrastructure for online RL. JTS designed and trained the RL token. ME built the intervention interface. AA and AE designed and built the gripper and robot hardware. CX, LK designed the system implementation, task suite and the experiments. SL, LK provided advice throughout the project. LK, CX, JTS, SL, ME worked on writing, illustrations and the video. 

\subsection{Additional experiment details}
\label{app:exp}


First, we collect a demonstration dataset on the target task; we then fine-tune the base VLA model and train the RL token for 2000 to 10000 gradient steps on the single task data. The VLA is then frozen during online RL training.

During online RL, we initialize the RL actor and critic from scratch with a two-layer MLP (hidden dimension 256) for the zip tie fastening, Ethernet, and charger insertion tasks. We use a larger network consisting of a three-layer MLP with hidden dimension 512 for the more challenging screw installation task. Both networks receive the RL token produced by the frozen base VLA model, proprioceptive position and velocity as input. The critic is trained with an ensemble of two Q functions following \citet{fujimoto2018td3} and we use the minimum of the two Q functions to calculate target values. The actor additionally takes in the reference action chunk produced by the VLA model, which is masked out at 50\% during training and always provided during inference. The actor is parameterized as a Gaussian policy with a small fixed standard deviation that outputs an action chunk $\mathbf{a}_{t:t+C-1} \in \mathbb{R}^{C \times d}$ with $C{=}10$ from the current observation. To increase sample efficiency, we subsample action chunks 2 control steps apart during training, so each second of data produces roughly 25 samples for the RL network. A sparse +1 reward is provided by the operator during training when the RL task has been completed. 

For the screw installation and zip tie fastening tasks, we first start the RL training in the critical-phase setting only. We then advance to a full task phase, first running the base model to complete the non-critical phase of the task, and switching to the RL policy when the critical phase is reached. This two-phase training strategy improves training efficiency while ensuring the RL policy is robust to the initial distributions induced by the base policy on the earlier part of the task. We report the policy performance after gathering about 5 hours of data.

\subsection{Additional experimental details for the baselines}
For all baseline methods, we use the same environment and action space setup as for our method -- policies execute in delta action space at 50\,Hz.
\label{app:baseline}

\textbf{PLD}: Following the original paper, we first pre-train the critic network on 50 base policy rollouts with Cal-QL~\cite{nakamoto2023cal} for better sample efficiency. We then proceed to the online RL stage.

\textbf{DSRL}: Following the original implementation, our implementation predicts a $(1, 32)$ dimensional latent action, which is repeated $50$ times in the first dimension to match the noise input space of our action-chunk VLA.

\textbf{HIL-SERL}: Following the original implementation, we initialize RLPD training with 20 episodes of demonstrations and provide interventions throughout training. However, it was not able to succeed in our setting because of the higher control frequency (50\,Hz) compared to the original system (10\,Hz) and the absence of an action-space bounding box to reduce exploration space. 

\textbf{DAgger}: We fine-tune our VLA with a mixture of demonstration data and the same set of intervention data collected during the online RL training. 

\end{document}